\documentclass[10pt,twocolumn,letterpaper]{article}

\usepackage[pagenumbers]{iccv} 

\newcommand{\red}[1]{{\color{red}#1}}
\newcommand{\todo}[1]{{\color{red}#1}}
\newcommand{\TODO}[1]{\textbf{\color{red}[TODO: #1]}}

\definecolor{iccvblue}{rgb}{0.21,0.49,0.74}
\usepackage[pagebackref,breaklinks,colorlinks,allcolors=iccvblue]{hyperref}
\usepackage{graphicx}
\graphicspath{{./imgs/}{./imgs/draw/}{./imgs/raw/}{./imgs/qualitative/}}
\usepackage{array}
\usepackage{booktabs}
\usepackage[table]{xcolor}

\def\paperID{*****} 
\def\confName{ICCV}
\def\confYear{2025}

\title{SmartFont: Dynamic Condition Allocation for Few-Shot Font Generation}

\author{Zian Yang\\
Fudan University\\
{\tt\small 23307140016@m.fudan.edu.cn}
\and
Zixin Wang\\
Fudan University\\
{\tt\small 23307130477@m.fudan.edu.cn}
}

\begin{document}
\maketitle

\begin{abstract}
Few-shot font generation simultaneously requires global structural completeness and fine-grained local style fidelity.
Existing methods usually either rely on global content-style modeling, which is robust but imperfectly disentangled, or emphasize component/local modeling, which captures fine details but relies heavily on local priors and reference coverage.
We argue that the key challenge is not merely to learn purer conditions, but to organize complementary yet biased global and local conditions through multi-level allocation during generation.
To this end, we propose SmartFont, a diffusion-based few-shot font generation framework that combines global content-style generation with weakly supervised local corrective experts.
The local branch performs semantic-spatial allocation by learning expert-wise local concepts and semantically meaningful spatial maps under weak component supervision, enabling fine-grained correction without requiring explicit component-conditioned inference.
On top of this, a denoising-state condition allocation module adaptively weights global content, global style, and local corrective feature across timesteps and injection blocks.
Extensive experiments show that SmartFont achieves better global-local balance, improves glyph quality and local detail fidelity.
\end{abstract}

\begin{figure}[t]
    \centering
    \includegraphics[page=5,width=\linewidth]{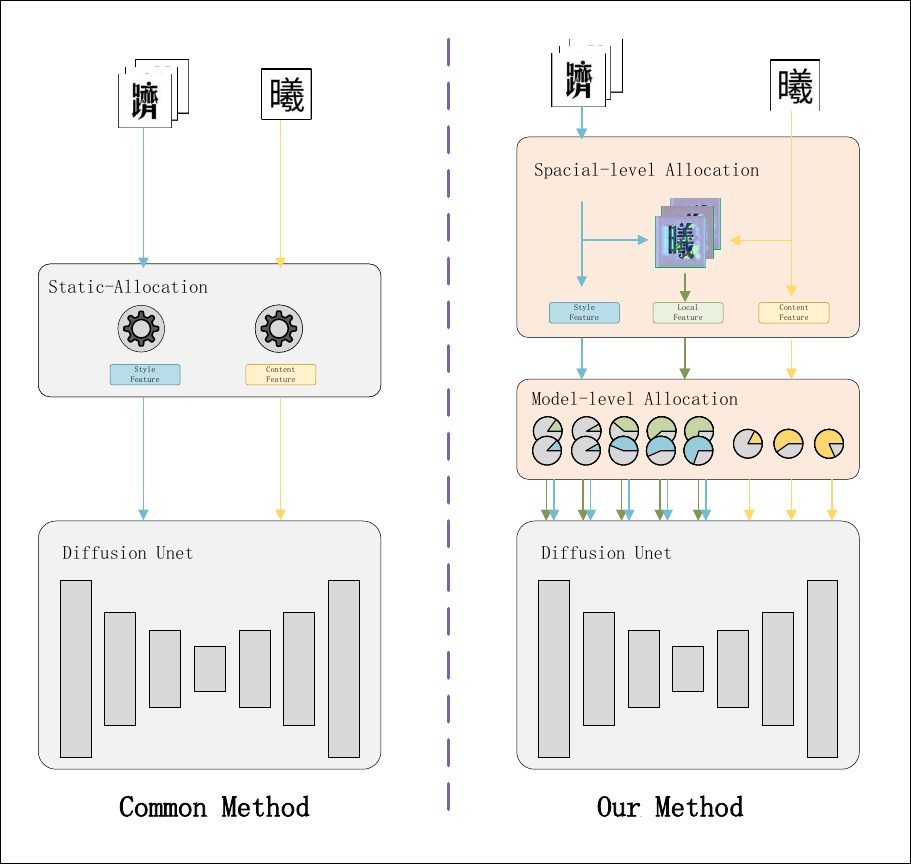}
    \caption{Comparison between conventional fixed-condition injection and our dynamic multi-level condition allocation. 
Left: common few-shot font Generation frameworks inject content and style conditions with fixed strengths at pre-defined denoising layers. 
Right: SmartFont augments the global diffusion backbone with a weakly supervised local expert branch, semantic-spatial allocation for localized correction, and denoising-state condition allocation that adaptively balances global content, global style, and local corrective feature across denoising states.}
    \label{intro}
\end{figure}

\begin{figure}[t]
    \centering

    \includegraphics[width=0.11\linewidth]{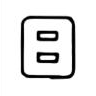}
    \hfill
    \includegraphics[width=0.11\linewidth]{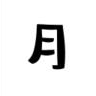}
    \hfill
    \includegraphics[width=0.11\linewidth]{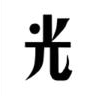}
    \hfill
    \includegraphics[width=0.11\linewidth]{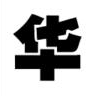}
    \hfill
    \includegraphics[width=0.11\linewidth]{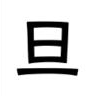}
    \hfill
    \includegraphics[width=0.11\linewidth]{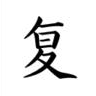}
    \hfill
    \includegraphics[width=0.11\linewidth]{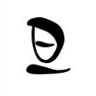}
    \hfill
    \includegraphics[width=0.11\linewidth]{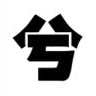}

    \vspace{0.3em}

    \includegraphics[width=0.11\linewidth]{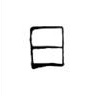}
    \hfill
    \includegraphics[width=0.11\linewidth]{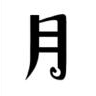}
    \hfill
    \includegraphics[width=0.11\linewidth]{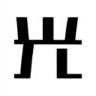}
    \hfill
    \includegraphics[width=0.11\linewidth]{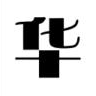}
    \hfill
    \includegraphics[width=0.11\linewidth]{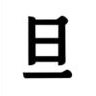}
    \hfill
    \includegraphics[width=0.11\linewidth]{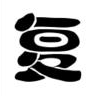}
    \hfill
    \includegraphics[width=0.11\linewidth]{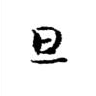}
    \hfill
    \includegraphics[width=0.11\linewidth]{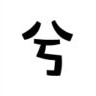}

    \vspace{0.3em}

    \includegraphics[width=0.11\linewidth]{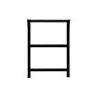}
    \hfill
    \includegraphics[width=0.11\linewidth]{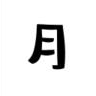}
    \hfill
    \includegraphics[width=0.11\linewidth]{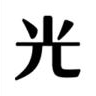}
    \hfill
    \includegraphics[width=0.11\linewidth]{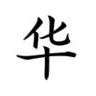}
    \hfill
    \includegraphics[width=0.11\linewidth]{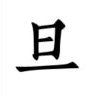}
    \hfill
    \includegraphics[width=0.11\linewidth]{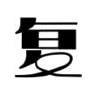}
    \hfill
    \includegraphics[width=0.11\linewidth]{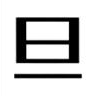}
    \hfill
    \includegraphics[width=0.11\linewidth]{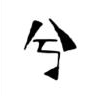}

    \caption{Qualitative results of SmartFont under diverse font styles. 
Each column corresponds to a different target font style, and the generated glyphs exhibit clear stylistic variation while preserving the target character structure. 
}
    \label{fig:eight_images}
\end{figure}

\section{Introduction}

Few-shot font generation aims to synthesize glyphs of a target font from only a limited set of reference characters, and has become an important problem for efficient font creation and stylized character generation.
A central challenge of this task is that the generated glyph should simultaneously preserve the global structure of the target character and faithfully inherit the fine-grained local style of the reference font.
This requirement is particularly demanding for complex ideographic characters, where even small stroke-level deviations may damage glyph readability or stylistic consistency, and recent studies\cite{yao2024vq,liu2023fonttransformer} still report clear difficulties in preserving complex structures and fine-grained stroke details under few-shot settings.

Existing few-shot font generation methods mainly follow two paradigms.
One line of work relies on global content--style modeling, CF-Font\cite{cf_font} and FontDiffuser\cite{fontdiffuser} encode content and style separately and then fuse them for glyph synthesis.
Such methods usually provide stronger global completeness and remain relatively robust when the reference set does not fully cover the target glyph.
However, such factorization is often only approximate in practice: even within disentanglement-based frameworks, the extracted content representation may remain suboptimal across fonts, and reference-character-specific structural bias may still leak into the generated result \cite{cf_font,chen2024iffont,he2024dsfont}.
Another line of work, including LF-Font\cite{lf_font_tpami}, MX-Font\cite{mx_font}, and later quantized local-style modeling methods\cite{pan2023vqlocal}, emphasizes component-level or localized style representations to better capture fine-grained font details.
Although such designs improve local stylization, they also make generation more sensitive to local prior quality, component definition, and reference coverage, and whole-glyph coherence can still be difficult when multiple local regions must be coordinated jointly \cite{lf_font_tpami,pan2023vqlocal}.
Although these methods improve local stylization, they are typically more sensitive to local priors, component definition, and reference coverage, and may struggle to maintain whole-glyph coherence when multiple local regions must be coordinated jointly.
Recent process-aware methods, such as Generate Like Experts\cite{gle_font} and FontAnimate\cite{fontanimate}, therefore reformulate font generation as a multi-stage or animated generation process rather than a single uniformly conditioned one-shot pipeline.

In this paper, we revisit few-shot font generation from the perspective of \emph{multi-level condition allocation}.
As shown in Fig.\ref{intro}, rather than viewing the problem solely as learning better disentangled content/style representations, we argue that the key challenge is to organize imperfect yet complementary conditions at multiple levels during generation.
At the spatial level, local style cues must be allocated to semantically meaningful experts and spatial regions, so that fine-grained correction can be applied where it is actually needed.
At the denoising-state level, global content, global style, and local corrective cues should not be treated as uniformly useful throughout the denoising process, since their relative utility varies across both injection blocks and denoising stages.
More broadly, recent conditional generation works such as CAN\cite{can}, SmartControl\cite{smartcontrol}, and Conditional Balance\cite{conditional_balance} suggest that different controls should not be used with a fixed globally shared strength, but should instead be coordinated adaptively across spatial locations, denoising stages, or conditions.

Based on this view, we propose \textbf{SmartFont}, a diffusion-based few-shot font generation framework that organizes global and local conditions through multi-level allocation.
SmartFont is built on a global diffusion backbone for basic content--style generation, and introduces a weakly supervised local expert branch to provide fine-grained local corrective cues.
The local branch performs semantic-spatial allocation by learning expert-wise local concepts under Hungarian-matching-based component supervision and by predicting semantically meaningful spatial maps for local correction.
On top of this, we design a denoising-state condition allocation module that adaptively weights global content, global style, and local corrective signals across both timesteps and injection blocks.
In this way, SmartFont coordinates where local correction should be applied and how strongly different conditions should contribute.

Fig.~\ref{fig:eight_images} presents representative results of SmartFont under diverse target font styles. 
Extensive experiments further show that SmartFont better balances global completeness and local fidelity than existing baselines.

Our main contributions are summarized as follows:
\begin{itemize}
    \item We formulate few-shot font generation from the perspective of \emph{multi-level condition allocation}, providing a unified view for coordinating global and local conditions.
    \item We propose SmartFont, a unified framework that combines a global diffusion backbone, a weakly supervised local expert branch for semantic-spatial allocation, and a denoising-state allocation module for adaptive multi-condition fusion.
    \item Extensive experiments demonstrate that SmartFont improves both global glyph quality and fine-grained local style fidelity, even under challenging few-shot and limited-coverage settings.
\end{itemize}

\section{Related Work}

\subsection{Diffusion Models}

Denoising Diffusion Probabilistic Models (DDPMs) \cite{ddpm} have established diffusion modeling as a standard framework for high-quality image generation by learning to reverse a gradual noising process, while Latent Diffusion Models (LDMs) \cite{ldm} further improve efficiency by performing diffusion in a compressed latent space.
Given a clean sample $x_0$, the forward process progressively perturbs it into noisy states $x_t$ according to
\begin{equation}
q(x_t \mid x_{t-1}) = \mathcal{N}\!\left(x_t; \sqrt{1-\beta_t}\,x_{t-1}, \beta_t I\right),
\end{equation}
and the reverse process is parameterized by a neural network $\epsilon_\theta$ that predicts the injected noise, or equivalently denoises the latent state.
A commonly used training objective is the standard noise-prediction loss
\begin{equation}
\mathcal{L}_{\mathrm{diff}}
=
\mathbb{E}_{x_0,\epsilon,t}
\left[
\left\|
\epsilon - \epsilon_\theta(x_t,t,\mathrm{cond})
\right\|_2^2
\right],
\end{equation}
where $\mathrm{cond}$ denotes external guidance such as content, style, or structural conditions.
In font generation, diffusion models provide a natural mechanism for injecting multiple conditions during denoising, which has motivated several recent few-shot font generation frameworks \cite{fontdiffuser,gle_font,fontanimate}.

\subsection{Few-shot Font Generation with Global Content--Style Modeling}
A major line of few-shot font generation methods follows the paradigm of global content--style disentanglement, where content and style are encoded separately and then fused for glyph synthesis.
Such methods typically provide a strong global generation prior and are relatively robust when reference coverage is limited, since they model style at the whole-glyph level rather than requiring explicit local matches.

Recent representative works further improve this paradigm from different aspects.
CF-Font\cite{cf_font} introduces content fusion and iterative style refinement to alleviate the sensitivity of content representations to the choice of representative font.
FontDiffuser\cite{fontdiffuser} formulates one-shot font generation in a diffusion framework and injects content information through multi-scale conditioning, further improving global structure preservation and stylization quality.
More recent variants also revisit this paradigm by strengthening the content condition itself or by replacing conventional disentanglement with richer semantic guidance, as exemplified by Conditional Font Generation with Content Pre-Train and Style Filter\cite{hong2025conditionalfont} and IF-Font\cite{chen2024iffont}.

Despite these advances, global content--style modeling still relies on only approximately disentangled conditions in practice.
Even within this paradigm, the extracted content feature may remain suboptimal across fonts, and style features may still carry residual character-structure bias, which can interfere with fine-grained local stylization \cite{cf_font,chen2024iffont,he2024dsfont}.
This limitation is especially visible when structurally different reference glyphs from the same font induce noticeably different local tendencies in the generated result.


\subsection{Component- and Local-Prior-Based Font Modeling}
Another line of work aims to improve few-shot font generation by introducing component-level or localized style modeling.
The motivation is that many font-specific characteristics are expressed through fine-grained local strokes, radicals, or substructures, which are difficult to capture using a single global style representation alone.

Representative methods in this direction include LF-Font\cite{lf_font_tpami}, MX-Font\cite{mx_font}, and Learning Component-Level and Inter-Class Glyph Representation for few-shot Font generation\cite{su2023componentglyph}.
LF-Font learns localized style representations instead of only universal style features, while MX-Font shows that weakly supervised local expert specialization can be effective under component-level supervision and Hungarian-based expert assignment.
Recent quantized local-style methods further reduce the reliance on manually predefined components by learning component-like local codes automatically\cite{pan2023vqlocal}.

Although local-prior methods are effective at improving local stylization, they are generally more sensitive to component definition, prior quality, and reference coverage.
Even when explicit manual components are avoided, the method still needs sufficiently informative local patterns to be exposed by the few-shot reference set.
Moreover, local fidelity does not automatically guarantee whole-glyph coherence, since a readable and stylistically consistent glyph depends not only on local correctness but also on the joint coordination of multiple parts \cite{lf_font_tpami,pan2023vqlocal}.

\subsection{Diffusion-Based and Process-Aware Font Generation}
Recent few-shot font generation works have increasingly explored diffusion-based and process-aware formulations.
Compared with earlier one-stage generation pipelines, these methods emphasize that different phases of glyph generation may require different kinds of guidance.

Generate Like Experts\cite{gle_font} models font generation as a multi-stage process consisting of structure construction, font generation, and refinement, and integrates this decomposition into a diffusion framework.
FontAnimate\cite{fontanimate} further reformulates font generation as an animated transition process and introduces condition alignment to improve both complex structure generation and detailed font style rendering.
Related calligraphy-oriented diffusion methods, such as DP-Font\cite{ijcai2024p0863}, additionally introduce stroke-order and physically inspired constraints to better handle complex character shapes and personalized writing styles.

These methods suggest that few-shot font generation should not be viewed as a uniformly conditioned one-shot fusion problem.
Instead, the usefulness of different conditions may vary across denoising stages or generation phases.

\subsection{Adaptive Condition Coordination in Generative Models}
Beyond font generation, recent conditional generation works have also shown that multiple conditions should not always be treated with fixed or uniformly shared strengths.
Instead, adaptive coordination of conditions can better resolve conflicts between different control sources.

CAN\cite{can} introduces condition-aware weight generation for controlled image generation, showing that generation quality can be improved when network responses are dynamically modulated by input conditions.
SmartControl\cite{smartcontrol} predicts spatially adaptive control scales to handle conflicts between text prompts and rough visual conditions, rather than enforcing a globally fixed control strength.
Conditional Balance\cite{conditional_balance} further analyzes style--content trade-offs under multiple conditions and shows that different denoising layers exhibit different sensitivities to different controls.

These works support a general principle that condition usage should be adaptive rather than static.
However, they are not designed for few-shot font generation, where the challenge additionally involves balancing global completeness, local fidelity, limited reference coverage, and imperfect content--style disentanglement.

\section{Observations and Motivations}

\begin{figure}[t]
    \centering
    \setlength{\tabcolsep}{0.5pt}
    \renewcommand{\arraystretch}{0.88}

    \newcommand{\imgw}{0.18\linewidth}
    \newcommand{\cellimg}[1]{%
        \includegraphics[width=\imgw]{#1}%
    }

    \begin{tabular}{@{}c@{\hspace{1pt}}c@{\hspace{1pt}}c@{\hspace{2pt}}c@{}}
        \scriptsize\textbf{Content} &
        \scriptsize\textbf{Reference} &
        \scriptsize\textbf{Result} &
        \scriptsize\textbf{GT} \\[-1pt]

        \cellimg{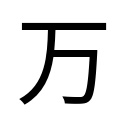} &
        \cellimg{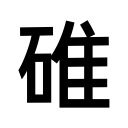} &
        \cellimg{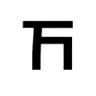} &
        \cellimg{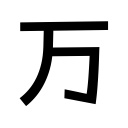} \\[-1pt]

        \cellimg{./imgs/raw/disentanglement/pair1/content.jpg} &
        \cellimg{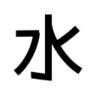} &
        \cellimg{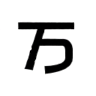} &
        \cellimg{./imgs/raw/disentanglement/pair1/gt.png} \\[-1pt]

        \cellimg{./imgs/raw/disentanglement/pair1/content.jpg} &
        \cellimg{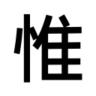} &
        \cellimg{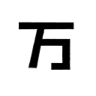} &
        \cellimg{./imgs/raw/disentanglement/pair1/gt.png} \\
    \end{tabular}

    \caption{Negative examples revealing imperfect disentanglement in global content--style modeling. 
With the target content fixed, changing only the reference character from the same font still leads to different structural tendencies in the generated results. 
In this case, reference glyphs dominated by horizontal/vertical strokes tend to make the target more rectilinear, while references with more slanted or curved strokes may induce undesired rounding or stroke bending.}
    \label{fig:disentanglement}
\end{figure}

\subsection{Imperfect disentanglement in global content-style modeling}
\label{sec:global_disentanglement_limit}

Global content--style modeling in few-shot font generation ideally assumes that content features preserve glyph identity while style features capture font-specific appearance independently of character identity. In practice, however, this factorization is only approximate.

Prior studies such as CF-Font\cite{cf_font}, IF-Font\cite{chen2024iffont}, and Few-Shot Font Generation by Learning Style Difference and Similarity\cite{he2024dsfont} collectively indicate that purely global content--style factorization remains imperfect in few-shot font generation.
Even within disentanglement-based frameworks, the extracted content representation itself may remain suboptimal across fonts, and residual character-structure bias may still leak into the style-related representation.
Under purely global modeling, such mixed conditions are especially harmful to fine-grained local stylization, because the style branch is expected to encode font-level appearance while remaining insensitive to reference-character-specific structure.

We observe the same phenomenon in our experiments. Even with a fixed target character, changing the reference character within the same font can still induce noticeably different structural tendencies in the generated results, as illustrated in Fig.\ref{fig:disentanglement}. These cases suggest that the global style branch may still carry non-negligible character-structure bias, rather than representing font style alone.

These observations suggest two practical limitations of global content--style modeling:
\begin{enumerate}
    \item Content and style are not fully disentangled in practice, so the resulting global conditions remain mixed and biased.
    \item Such imperfect disentanglement is particularly detrimental to fine-grained stroke details, often causing local distortions and reduced style fidelity.
\end{enumerate}

\subsection{Limitations of component-dependent local modeling}
\label{sec:local_modeling_limit}
Many few-shot font generation methods introduce component-wise or localized style modeling to improve local stylistic fidelity \cite{lf_font_tpami,su2023componentglyph,pan2023vqlocal}.
Such designs are well motivated for glyph-rich scripts, because many font-specific characteristics are expressed through fine-grained local strokes, radicals, or substructures that are difficult to capture using a single global style representation alone.

However, this advantage also comes with a stronger dependence on local priors and reference coverage.
In practice, local units such as components or substructures are not always stable visual entities across fonts: even the same nominal component can exhibit location-dependent profiles, and mismatched candidate locations may lead to inconsistent local stylization \cite{zhao2024calligraphy}.
Moreover, pre-defined local correspondences are not always a convenient or scalable prior, so some methods instead learn component-like local codes automatically rather than relying on fixed component definitions \cite{pan2023vqlocal}.
Under few-shot settings, the reference set may also fail to expose all locally relevant patterns needed by the target glyph \cite{lf_font_tpami}.
As a result, the effectiveness of component-dependent methods becomes sensitive to both prior definition and reference coverage.

Another limitation is that local correctness does not automatically imply whole-glyph quality. A glyph is not merely a collection of independently stylized parts; its final appearance also depends on how multiple local regions are composed into a coherent structure. When the modeling focus is overly localized, the generated result may preserve certain local style cues while still suffering from mismatched proportions, awkward part interactions, or unstable overall structure. Recent studies including FontAnimate\cite{fontanimate} and Generate Like Experts\cite{gle_font} likewise suggest that local detail modeling alone is insufficient for preserving accurate complex structures and coherent glyph composition in few-shot font generation.

Taken together, these observations suggest two practical limitations of component-dependent local modeling:
\begin{enumerate}
    \item Its effectiveness is highly sensitive to local prior definition and reference coverage, especially when the target glyph requires weakly exposed or unseen local patterns.
    \item Its local focus does not guarantee whole-glyph coherence, and may still lead to structural instability.
\end{enumerate}

\subsection{Block-wise and timestep-wise sensitivity to content and style conditions}
\label{sec:statewise_condition_sensitivity}
\begin{figure}[t]
    \centering

    \begin{subfigure}[t]{0.49\linewidth}
        \centering
        \includegraphics[page=7,width=\linewidth]{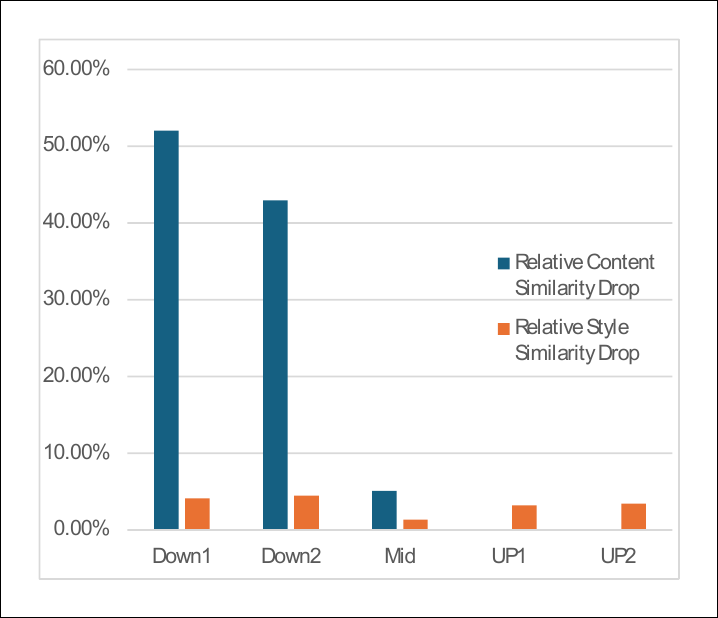}
        \caption{Layer observation.}
        \label{fig:observation_layer}
    \end{subfigure}
    \hfill
    \begin{subfigure}[t]{0.49\linewidth}
        \centering
        \includegraphics[page=6,width=\linewidth]{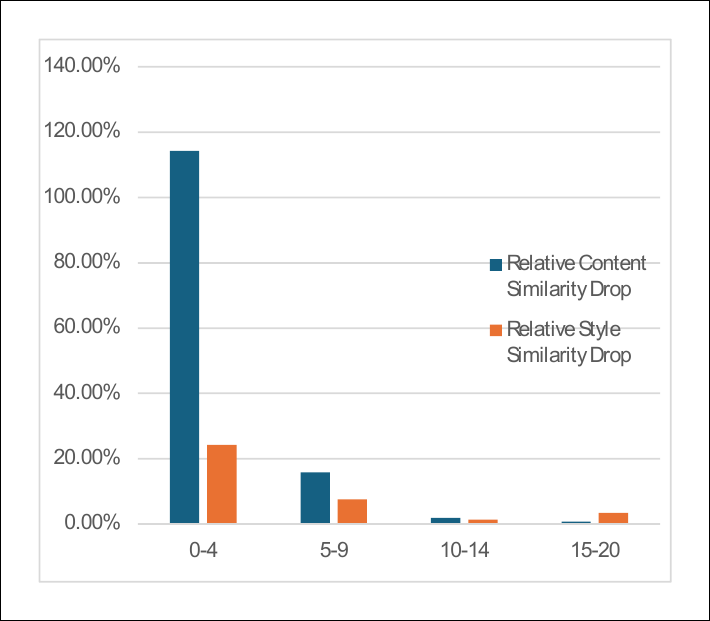}
        \caption{Timestep observation.}
        \label{fig:observation_timestep}
    \end{subfigure}

\caption{Observations of state-dependent sensitivity to content and style conditions from two perspectives. 
Left: relative similarity drop caused by attenuating content or style conditions at different injection blocks. 
Right: relative similarity drop caused by attenuating the same conditions over different timestep ranges. 
}
\label{fig:observation_combined}
\end{figure}

\begin{figure}[t]
    \centering
    \setlength{\tabcolsep}{1pt}
    \renewcommand{\arraystretch}{0.95}

    \newcommand{\cellimg}[1]{%
        \includegraphics[width=\linewidth]{#1}%
    }

    {\scriptsize\textbf{(a) Block-wise ablation}}\\[-1mm]
    \begin{tabular}{@{}>{\centering\arraybackslash}m{0.13\linewidth}
                    *{6}{>{\centering\arraybackslash}m{0.11\linewidth}}@{}}
        & \scriptsize\textbf{Down1}
        & \scriptsize\textbf{Down2}
        & \scriptsize\textbf{Mid}
        & \scriptsize\textbf{Up1}
        & \scriptsize\textbf{Up2}
        & \scriptsize\textbf{Normal} \\

        \scriptsize\textbf{Content}
        & \cellimg{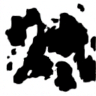}
        & \cellimg{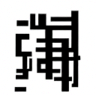}
        & \cellimg{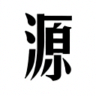}
        & \mbox{}
        & \mbox{}
        & \cellimg{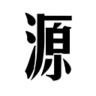} \\

        \scriptsize\textbf{Style}
        & \cellimg{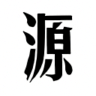}
        & \cellimg{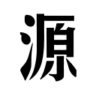}
        & \cellimg{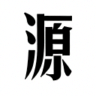}
        & \cellimg{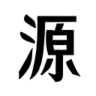}
        & \cellimg{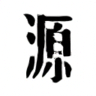}
        & \cellimg{./imgs/raw/timestep/normal.png} \\
    \end{tabular}

    \vspace{2mm}

    {\scriptsize\textbf{(b) Timestep-wise ablation}}\\[-1mm]
    \begin{tabular}{@{}>{\centering\arraybackslash}m{0.13\linewidth}
                    *{5}{>{\centering\arraybackslash}m{0.13\linewidth}}@{}}
        & \scriptsize\textbf{0--4}
        & \scriptsize\textbf{5--9}
        & \scriptsize\textbf{10--14}
        & \scriptsize\textbf{15--20}
        & \scriptsize\textbf{Normal} \\

        \scriptsize\textbf{Content}
        & \cellimg{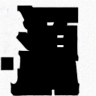}
        & \cellimg{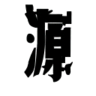}
        & \cellimg{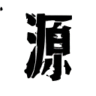}
        & \cellimg{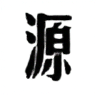}
        & \cellimg{./imgs/raw/timestep/normal.png} \\

        \scriptsize\textbf{Style}
        & \cellimg{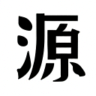}
        & \cellimg{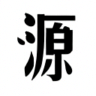}
        & \cellimg{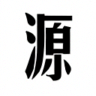}
        & \cellimg{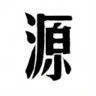}
        & \cellimg{./imgs/raw/timestep/normal.png} \\
    \end{tabular}

    \caption{Qualitative ablation results from block-wise and timestep-wise perspectives.
    (a) Each column shows the generated glyph when the corresponding content or style condition is attenuated only at a specific injection block, while the other conditions remain unchanged; the last column shows the normal result.
    (b) Each column shows the result when the corresponding condition is attenuated only within a specific timestep range.
    }
    \label{fig:qual_observation}
\end{figure}

Recent conditional generation studies suggest that different controls may exert non-uniform influence across network states, rather than being used with a fixed globally shared strength \cite{can,smartcontrol,conditional_balance}.
Motivated by this principle, we ask whether the denoising network in few-shot font generation likewise responds equally to content and style conditions across different states.
To answer this, we analyze the sensitivity of the denoising process to content and style conditions at different injection blocks and timestep ranges.

Starting from a standard dual-condition diffusion backbone, we attenuate one condition at a given injection block or timestep range while keeping the others unchanged, and measure the resulting degradation from both visual quality and feature similarity. Specifically, we compare generated glyphs with the ground truth in content/style feature space using cosine similarity and $\ell_2$ distance, and use their averaged percentage degradation as a practical proxy for condition sensitivity drop.

Our analysis reveals clear \emph{block-wise heterogeneity}. Attenuating the same condition at different injection blocks leads to noticeably different degrees of degradation in feature similarity and visual quality, as also reflected by the qualitative results in Fig.\ref{fig:observation_layer} and Fig.\ref{fig:qual_observation}. This suggests that condition sensitivity is not uniform across blocks, but depends on where the condition is consumed within the denoising network.

A similar trend appears along the temporal axis. When the denoising process is divided into multiple timestep ranges, attenuating the same condition in different ranges also leads to different degrees of degradation, as shown in Fig.\ref{fig:observation_timestep} and Fig.\ref{fig:qual_observation}. This indicates that condition sensitivity is likewise non-uniform over time, and that the usefulness of a condition depends not only on \emph{what} it represents, but also on \emph{when} it is injected.

Together, these observations suggest the following implications.
\begin{enumerate}
    \item Different injection blocks exhibit different relative sensitivities to content and style conditions.
    \item Different timestep ranges also rely on content and style conditions to different extents.
    \item Therefore, condition fusion should be treated as a state-dependent allocation problem, rather than a static globally shared design.
\end{enumerate}

\begin{figure*}[t]
    \centering
    \includegraphics[page=1,width=\linewidth]{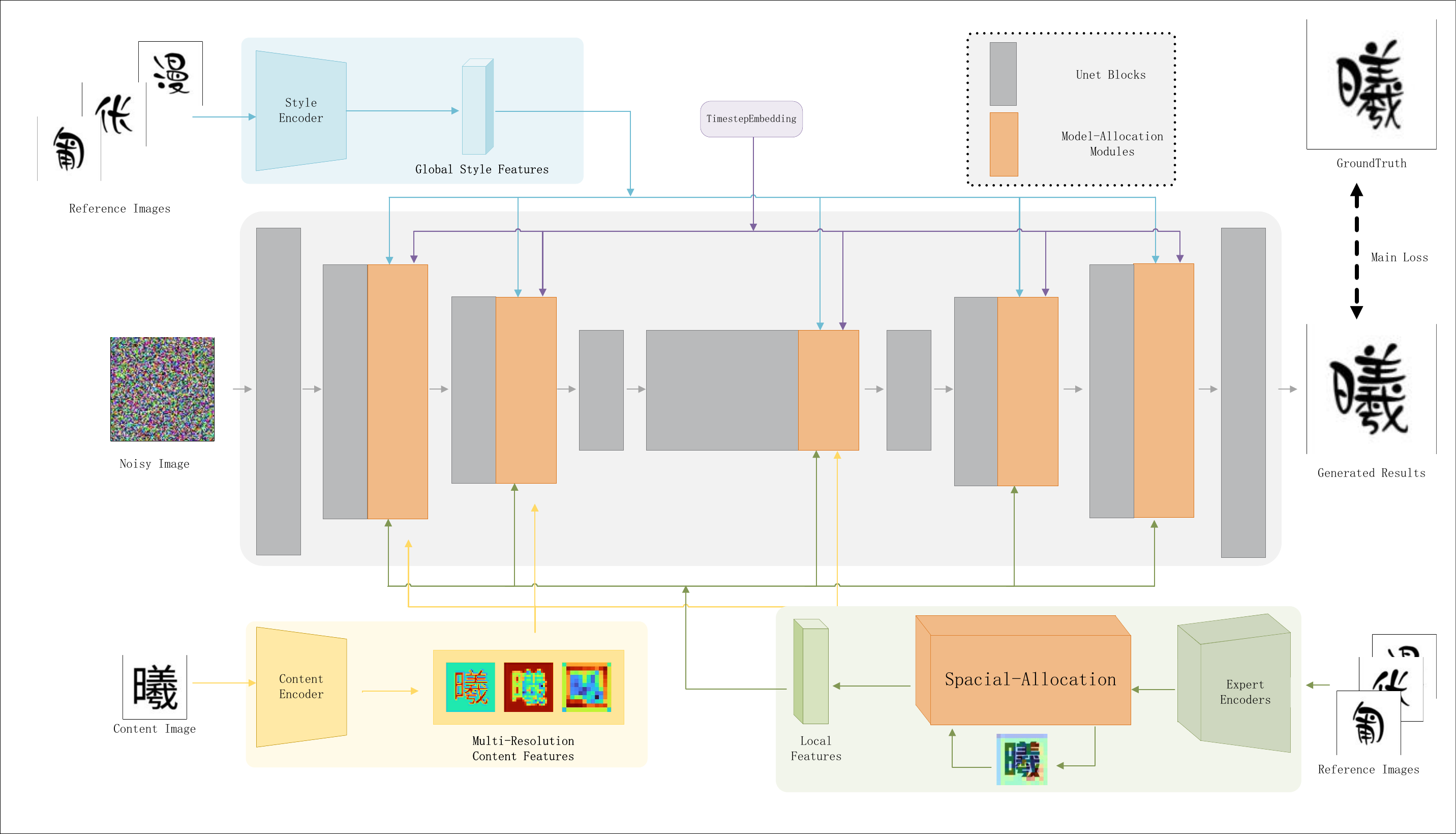}
\caption{Overview of SmartFont. 
Built on a global diffusion font generation backbone, SmartFont coordinates three complementary conditions---global content, global style, and local corrective cues---through a unified multi-level allocation framework. 
The weakly supervised local expert branch performs semantic-spatial allocation by extracting expert-wise local factors and predicting spatial weight maps for localized correction, while the denoising-state allocation modules perform allocation by adaptively weighting different conditions across injection blocks and timesteps. 
These allocated conditions are injected into the U-Net denoiser to generate the target glyph under supervision from the ground truth.
    \label{mainstructure}
}
\end{figure*}

\section{Method}

\subsection{Overview}
\label{sec:overview}

Based on the observations in \cref{sec:global_disentanglement_limit,sec:local_modeling_limit,sec:statewise_condition_sensitivity}, we reformulate few-shot font generation as a \emph{multi-level condition allocation} problem.
Rather than relying on a fixed globally shared fusion rule, SmartFont organizes imperfect yet complementary conditions at two levels during generation: \emph{semantic-spatial allocation}, which determines which local concepts should contribute and where they should be applied, and \emph{denoising-state allocation}, which determines how all global and local conditions should be weighted across different blocks and timestep ranges.
To this end, SmartFont introduces three complementary condition sources, namely a global content condition, a global style condition, and a local corrective condition, and coordinates them through a unified allocation framework.

As shown in Fig.~\ref{mainstructure}, SmartFont is built on a diffusion font generation backbone for basic global content--style generation.
To complement this global pathway, we introduce a weakly supervised multi-expert local branch that extracts expert-wise local style factors together with local assignment cues.
The assignment cues are further combined with content features to predict semantically meaningful spatial weight maps, enabling different experts to specialize in different local concepts and regions.
Building on these conditions, SmartFont performs two forms of adaptive allocation: semantic-spatial allocation through the local expert branch, and denoising-state allocation through a dedicated condition weighting module.
The former is built on weakly supervised expert specialization inspired by MX-Font~\cite{mx_font}, with Hungarian-matching-based supervision redesigned for our diffusion setting, while the latter dynamically balances global content, global style, and local corrective conditions across blocks and timesteps during few-shot font generation.
The global diffusion backbone, the weakly supervised local expert branch, the denoising-state condition allocation module, and the training objectives are detailed in \cref{sec:backbone,sec:local_expert,sec:condition_weighting,sec:training_objectives}, respectively.

\subsection{Global diffusion backbone}
\label{sec:backbone}

We instantiate SmartFont on a global conditional diffusion backbone that provides the basic global content--style generation pathway. This backbone follows the conditional diffusion paradigm of FontDiffuser\cite{fontdiffuser}, where a source glyph provides content guidance and a reference glyph provides style guidance to the denoising network. Specifically, we retain a U-Net denoiser together with the corresponding content encoder, style encoder, and multi-point global condition injection pathway, which jointly serve as the global generation branch of our framework.

Unlike FontDiffuser, however, we do not adopt its Style Contrastive Refinement (SCR) module. Our goal is not to further improve the global style branch through an additional style representation learning objective, but to study how imperfect global conditions should be complemented by local corrective cues and dynamically allocated across denoising states. Therefore, we use the diffusion backbone as a standard global content--style foundation, upon which our main contributions are built.


\begin{figure}[t]
    \centering
    \includegraphics[page=2,width=\linewidth]{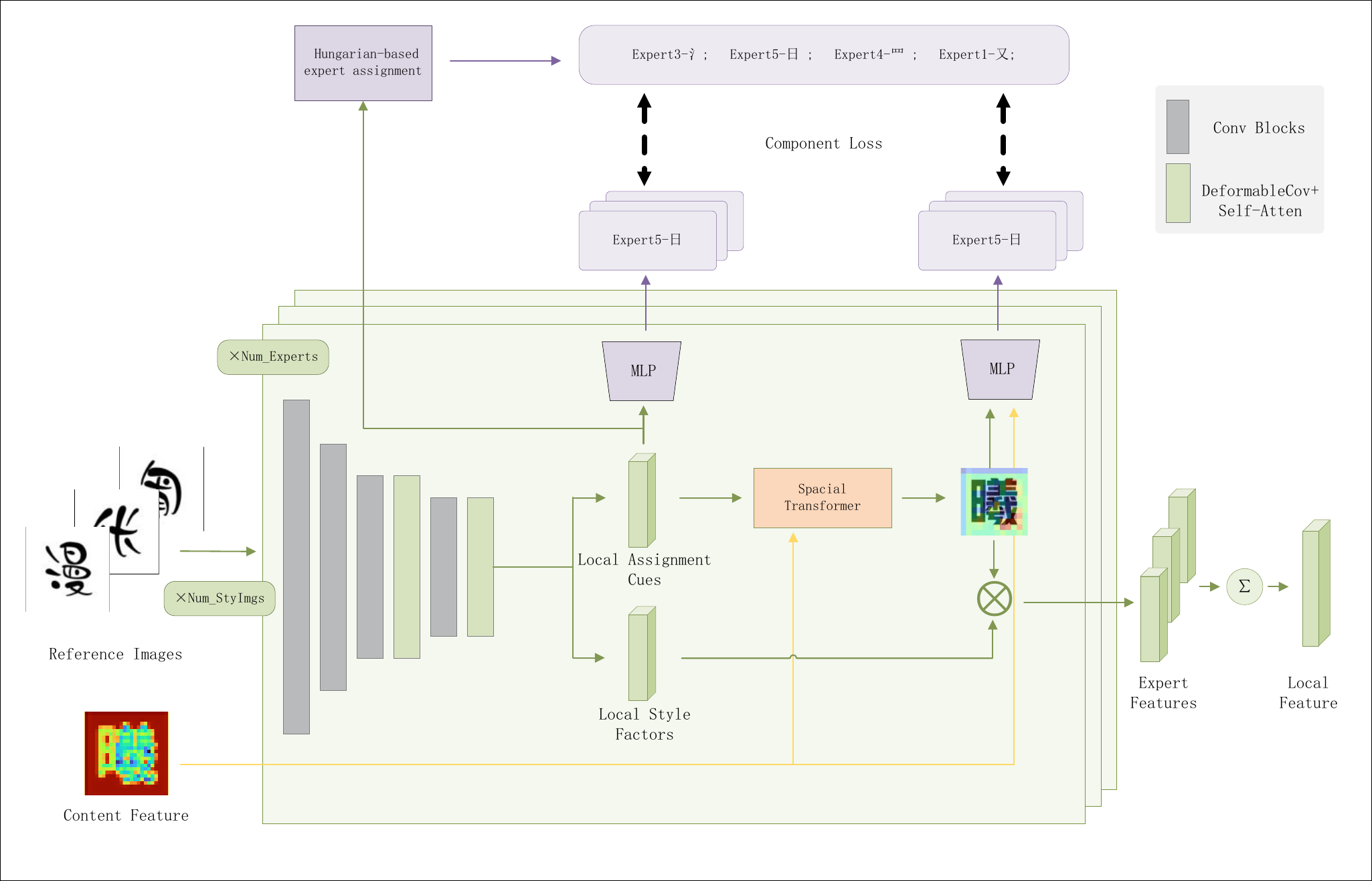}
\caption{Architecture of the weakly supervised local expert branch.
A multi-expert local encoder produces expert-wise local features, which are split into local style factors and local assignment cues.
Hungarian-matched weak supervision drives expert specialization, while content-aware spatial maps modulate and aggregate the local factors into a local corrective feature.}
    \label{local_allocation}
\end{figure}

\subsection{Weakly-supervised local expert branch}
\label{sec:local_expert}

The global diffusion backbone provides the basic content--style generation pathway, but remains limited in capturing fine-grained local stroke variation. To complement this global pathway without reverting to explicit component-conditioned generation, we introduce a weakly supervised local expert branch for local corrective modeling. 
Inspired by the multiple localized experts paradigm of MX-Font\cite{mx_font}, this branch adopts multi-expert learning together with weak component supervision and Hungarian-based assignment, while being redesigned as a local corrective module rather than a standalone generator. 
Concretely, it extracts expert-wise local features, splits them into local style factors and local assignment cues, and converts them into spatially localized corrective features through semantic-aware spatial allocation.

\subsubsection{Expert-wise local feature extraction}
\label{sec:expert_feature_extraction}

We employ a multi-expert local encoder to extract expert-wise local features from the input glyph. As shown in Fig.~\ref{local_allocation}, the encoder is built from stacked convolution, self-attention, and deformable convolution modules. Here, the deformable convolution is introduced to better accommodate the highly non-uniform and elongated nature of font strokes. Each expert is expected to capture a different local concept, so that the branch can model diverse local variations that are difficult to express with a single global condition.

Let
\begin{equation}
\{e_i\}_{i=1}^{K}=E_{\mathrm{loc}}(x_{\mathrm{ref}}),
\qquad
[s_i,\; q_i]=\mathrm{Split}(e_i),
\end{equation}
where $e_i$ denotes the feature of the $i$-th expert, $s_i$ denotes the corresponding local style factor, and $q_i$ denotes the local assignment cue. The local style factors encode \emph{what} local appearance may be injected, while the local assignment cues encode \emph{which} local concept an expert tends to represent and later provide the basis for \emph{where} such correction should be applied.

Notably, $s_i$ and $q_i$ are derived from shared expert features and are only separated at the final stage, rather than being strictly disentangled throughout the encoder. This design keeps expert learning grounded in common local evidence, while leaving the subsequent spatial allocation module to further determine where each expert-specific correction should be applied.

\begin{figure}[t]
    \centering
    \includegraphics[page=3,width=\linewidth]{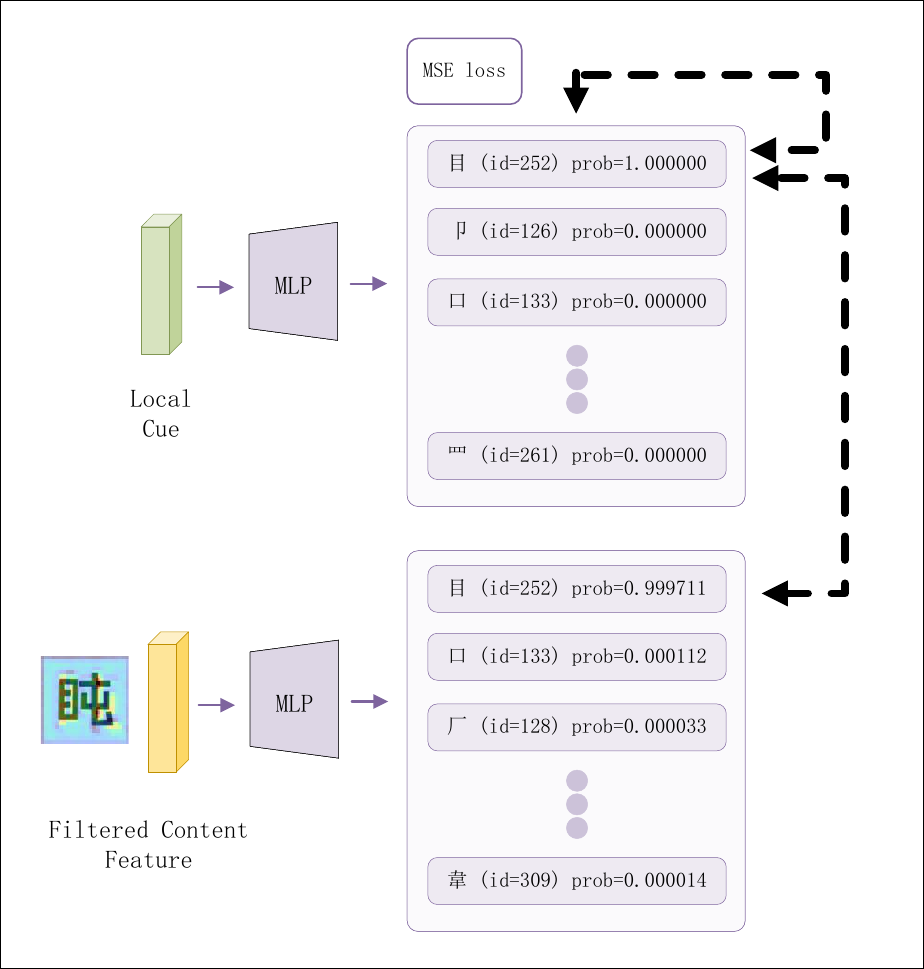}
\caption{Hungarian-based expert assignment under weak component supervision.
Each expert predicts component logits from its local assignment cue over the image-level component set.
These logits form the expert--component score matrix, from which Hungarian matching produces a one-to-one assignment, which is then used to supervise the cue branch and the spatial-map branch, coupling expert specialization with spatial localization.}
    \label{Hungarian_match}
\end{figure}

\subsubsection{Hungarian-based expert assignment}
\label{sec:hungarian_assignment}

To prevent different experts from collapsing to similar local patterns, we impose weak component supervision on the assignment-cue branch, following the general idea of MX-Font \cite{mx_font}. Instead of predefining a fixed correspondence between expert indices and component categories, we infer an expert--component matching automatically during training, as illustrated in Fig.~\ref{Hungarian_match}.

For each expert $i$, let $q_i$ denote its local assignment cue, and let $\phi(q_i)\in\mathbb{R}^{|\mathcal U|}$ be the predicted logits over the image-level component set $\mathcal U$. We define the expert--component matching score as
\begin{equation}
S_{ij} = \phi(q_i)_j ,
\end{equation}
and compute the one-to-one assignment matrix $A=[a_{ij}]$ by Hungarian matching:
\begin{equation}
A^\star
=
\arg\max_{A\in\Pi(\mathcal U)}
\sum_{i=1}^{K}\sum_{j\in\mathcal U} a_{ij} S_{ij},
\end{equation}
where $\Pi(\mathcal U)$ denotes the feasible assignment set.

The resulting assignment supervises the expert-specific assignment cues and is further shared with the spatial-map branch, with the corresponding auxiliary objectives defined in Sec.~\ref{sec:training_objectives}. In this way, expert specialization and spatial localization are learned under the same weak semantic constraint, rather than emerging as unconstrained attention patterns; typical degeneration cases without such constraints are discussed later in Sec.~\ref{sec:vis_spatial} and visualized in Fig.~\ref{fig:naive_gate_failure}.

\subsubsection{Semantic-Spatial allocation from assignment cues}
\label{sec:spatial_allocation}

This branch performs \emph{semantic-aware spatial allocation} for local correction, namely, determining where each expert-specific correction should be applied on the target glyph. Let $F\in\mathbb{R}^{C\times H\times W}$ denote the content feature map used as the structural anchor. For each expert, we combine $F$ with the corresponding local assignment cue $q_i$ to predict expert-wise spatial logits:
\begin{equation}
L_i = T_{\mathrm{sp}}(F, q_i),
\qquad
G_i = \sigma(L_i),
\end{equation}
where $T_{\mathrm{sp}}(\cdot)$ denotes the spatial transformer and $G_i$ is the resulting spatial gate map.

The predicted map is first used for local corrective injection. Specifically, the $i$-th expert contributes
\begin{equation}
v_i = s_i \odot G_i,
\qquad
f_{\mathrm{loc}} = \sum_{i=1}^{K} v_i,
\end{equation}
where $\odot$ denotes element-wise multiplication. In this way, $s_i$ provides expert-wise local appearance information, while $G_i$ determines its spatial contribution on the target glyph.

The same spatial logits are also used for semantic readout on the content feature map:
\begin{align}
P_i(u,v) &= \frac{\exp(L_i(u,v))}{\sum_{u',v'}\exp(L_i(u',v'))}, \\
z_{i,c} &= \sum_{u,v} P_i(u,v)\,F_{cuv}, \\
m_i &= \frac{1}{HW}\sum_{u,v} G_i(u,v), \\
h_i &= [z_i; m_i].
\end{align}
The readout feature $h_i$ is then fed to the map classification head defined in Sec.~\ref{sec:training_objectives}. Therefore, the Hungarian-matched assignment supervises not only the assignment-cue branch but also the spatial-map branch, tying expert specialization and spatial localization to the same weak semantic constraint.

We use the content feature $F$ rather than the evolving noisy latent as the map-prediction anchor because it already preserves the global layout of the target glyph and is more stable for low-resolution region localization. As a result, the learned maps are encouraged to highlight semantically relevant local regions rather than degenerating into trivial responses. 
In this way, the local expert branch acts not as an independent component-conditioned generator, but as a weakly supervised local corrective module that complements the global diffusion backbone with localized detail cues.

\begin{figure}[t]
    \centering
    \includegraphics[page=4,width=\linewidth]{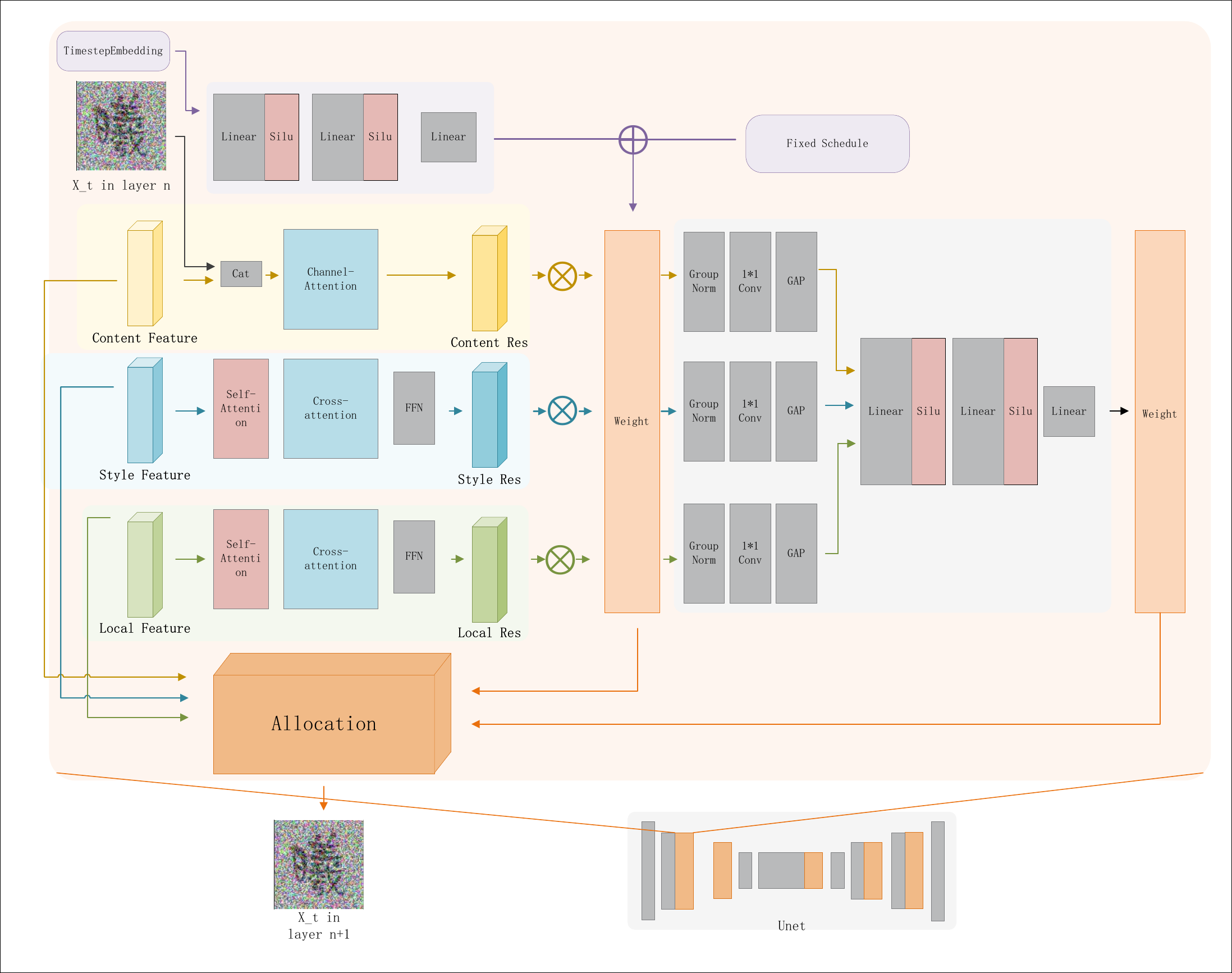}
    \caption{Detailed architecture of the denoising-state allocation module. 
    A timestep-aware base allocation is first predicted from the timestep embedding, and then refined at each injection layer using the current denoising feature and the condition-specific residual responses. 
    The resulting weights modulate the global content, global style, and local corrective pathways before fusion.}
    \label{model_allocation}
\end{figure}

\subsection{Denoising-state condition allocation and weighting}
\label{sec:condition_weighting}

The denoising-state condition allocation module determines how different conditions should be weighted at different blocks and timesteps.
Motivated by both our sensitivity analysis in \cref{sec:statewise_condition_sensitivity} and recent adaptive condition-coordination strategies in conditional generation \cite{can,smartcontrol,conditional_balance}, we introduce a unified allocation-and-weighting module to adaptively balance global content, global style, and local corrective cues during denoising.

As shown in Fig.\ref{model_allocation}, it first produces a timestep-aware base allocation, and then refines it using the current block feature. As a result, the final condition coefficients depend on both the denoising stage and the feature state of the current injection block.

\subsubsection{Formulation}
\label{sec:weighting_motivation}

At each injection layer, the available conditional responses are not fused with fixed coefficients. Instead, we predict layer- and timestep-dependent condition weights to modulate their relative contribution before injection, thereby formulating condition fusion as a denoising-state allocation problem.

For a given injection layer $l$ at timestep $t$, let $h^{(l,t)}$ denote the current denoising feature, and let
$r_c^{(l,t)}$, $r_s^{(l,t)}$, and $r_l^{(l,t)}$
denote the conditional responses from the global content, global style, and local corrective pathways, respectively. The weighted fusion is defined as
\begin{equation}
\tilde{h}^{(l,t)}
=
h^{(l,t)}
+
w_c^{(l,t)} r_c^{(l,t)}
+
w_s^{(l,t)} r_s^{(l,t)}
+
w_l^{(l,t)} r_l^{(l,t)},
\end{equation}
where $\mathbf{w}^{(l,t)}=[w_c^{(l,t)},w_s^{(l,t)},w_l^{(l,t)}]$ is the condition-weight vector predicted by the proposed denoising-state allocation module.

\subsubsection{Timestep-aware base weighting}
\label{sec:timestep_weighting}

We first introduce a timestep-aware base weighting to provide a coarse prior for condition usage along the denoising trajectory. For each injection layer $l$ at timestep $t$, we define a base condition-weight vector
$\mathbf{w}^{\mathrm{base}}_{l,t}\in\mathbb{R}^{3}$ over the global content, global style, and local corrective pathways as
\begin{equation}
\mathbf{w}_{\mathrm{base}}^{(l,t)}
=
\mathbf{s}^{(l)}(t) + f_{\mathrm{temp}}^{(l)}(e_t),
\end{equation}
where $\mathbf{s}_{l}(t)$ is a layer-specific empirical schedule, $e_t$ is the timestep embedding, and $f_{\mathrm{temp}}^{(l)}$ is a lightweight MLP that predicts a residual adjustment for the three conditions at layer $l$. In this way, the empirical schedule serves only as a coarse temporal prior, while the learnable residual allows different layers to adapt their condition preference across denoising stages.

\subsubsection{Layer-adaptive weighting refinement}
\label{sec:block_weighting}

Timestep-aware base weighting captures coarse stage-level tendencies, but the same timestep may still require different condition balances at different injection layers. We therefore further refine the base allocation using the current denoising state at each layer. As illustrated in Fig.~\ref{model_allocation}, the refinement is conditioned not only on the current denoising feature, but also on the timestep-weighted conditional residuals from the global content, global style, and local corrective pathways.

Specifically, for each injection layer $l$ at timestep $t$, let
$r_c^{(l,t)}$, $r_s^{(l,t)}$, and $r_l^{(l,t)}$
denote the conditional responses from the global content, global style, and local corrective pathways, respectively. We first compute the timestep-weighted residuals
\begin{equation}
\hat{r}_k^{(l,t)}
=
w_{k,\mathrm{base}}^{(l,t)} \, r_k^{(l,t)},
\qquad
k \in \{c,s,l\},
\end{equation}
where $\mathbf{w}_{\mathrm{base}}^{(l,t)}
=
\left[
w_{c,\mathrm{base}}^{(l,t)},
w_{s,\mathrm{base}}^{(l,t)},
w_{l,\mathrm{base}}^{(l,t)}
\right]$
is the timestep-aware base weight vector.

We then predict a layer-adaptive multiplicative refinement factor from the current denoising feature and the three timestep-weighted residuals:
\begin{equation}
\mathbf{g}^{(l,t)}
=
f_{\mathrm{gate}}^{(l)}
\!\left(
h^{(l,t)},
\hat{r}_c^{(l,t)},
\hat{r}_s^{(l,t)},
\hat{r}_l^{(l,t)}
\right),
\end{equation}
and obtain the final condition weights by element-wise multiplication
\begin{equation}
\mathbf{w}^{(l,t)}
=
\mathbf{w}_{\mathrm{base}}^{(l,t)}
\odot
\mathbf{g}^{(l,t)},
\end{equation}
where $\odot$ denotes element-wise multiplication and
$\mathbf{w}^{(l,t)}=[w_c^{(l,t)},w_s^{(l,t)},w_l^{(l,t)}]$.
In this way, the refinement becomes directly aware of the candidate changes induced by different conditions at the current layer state, rather than relying on the raw denoising feature alone. The resulting weights are then used to modulate the corresponding conditional responses before fusion.

\subsection{Training objectives}
\label{sec:training_objectives}

Our training objective consists of a main generation objective, a weak semantic supervision term for expert specialization and map learning, and a set of relative map activation priors for stabilizing semantic-spatial allocation.

\paragraph{Main generation objective.}
We optimize the diffusion backbone with the standard denoising loss and a perceptual content loss:
\begin{equation}
\mathcal{L}_{\mathrm{main}}
=
\mathcal{L}_{\mathrm{diff}}
+
\lambda_{cp}\mathcal{L}_{cp}
+
\lambda_{off}\mathcal{L}_{off},
\end{equation}
where $\mathcal{L}_{\mathrm{diff}}$ is the standard MSE diffusion loss, $\mathcal{L}_{cp}$ is the perceptual content loss, and $\mathcal{L}_{off}$ is the offset regularization when the deformation branch is retained.

\paragraph{Weak semantic supervision.}
Let $A=[a_{ij}]$ denote the expert--component assignment obtained by Hungarian matching, where $a_{ij}=1$ indicates that expert $i$ is matched to component $j$. This assignment provides weak supervision for both the assignment-cue branch and the spatial-map branch.

Each expert predicts a component distribution from its local assignment cue. Let $p_i^{\mathrm{cue}}$ denote the cue-based prediction of expert $i$. The corresponding cue classification loss is
\begin{equation}
\mathcal{L}_{\mathrm{cls}}^{\mathrm{cue}}
=
\sum_{i=1}^{K}\sum_{j\in\mathcal U}
a_{ij}\,\mathrm{CE}(p_i^{\mathrm{cue}}, j),
\end{equation}
where $\mathcal U$ is the image-level component set of the current glyph.

For the spatial-map branch, we directly reuse the map readout feature $h_i$ defined in Sec.~\ref{sec:spatial_allocation}, and feed it to a map classifier:
\begin{equation}
p_i^{\mathrm{map}}=\mathrm{MLP}(h_i).
\end{equation}
The corresponding map classification loss is
\begin{equation}
\mathcal{L}_{\mathrm{cls}}^{\mathrm{map}}
=
\sum_{i=1}^{K}\sum_{j\in\mathcal U}
a_{ij}\,\mathrm{CE}(p_i^{\mathrm{map}}, j).
\end{equation}
We combine the two terms as
\begin{equation}
\mathcal{L}_{\mathrm{sem}}
=
\lambda_{\mathrm{cue}} \mathcal{L}_{\mathrm{cls}}^{\mathrm{cue}}
+
\lambda_{\mathrm{map}} \mathcal{L}_{\mathrm{cls}}^{\mathrm{map}}.
\end{equation}

\paragraph{Relative map activation priors.}
Besides semantic classification, we further regularize the relative activation strength of the predicted maps. Let
\begin{equation}
\bar{G}_i=\mathrm{mean}(G_i)
\end{equation}
denote the mean activation of the $i$-th gate map, where $G_i$ is defined in Sec.~\ref{sec:spatial_allocation}. Let $\mathcal S_{\mathrm{sel}}$ be the set of experts selected by Hungarian assignment, and let $\mathcal S_{\mathrm{ovl}}\subseteq\mathcal S_{\mathrm{sel}}$ denote the subset whose matched components overlap with those actually present in the target glyph.

We first encourage selected maps to respond more strongly than the overall average:
\begin{equation}
\mathcal{L}_{\mathrm{map}}^{\mathrm{sel}}
=
-\log
\frac{
\frac{1}{|\mathcal S_{\mathrm{sel}}|}
\sum_{i\in\mathcal S_{\mathrm{sel}}}\bar{G}_i + \epsilon
}{
\frac{1}{K}\sum_{i=1}^{K}\bar{G}_i + \epsilon
}.
\end{equation}
We further encourage overlapping selected maps to respond more strongly than the remaining selected ones:
\begin{equation}
\mathcal{L}_{\mathrm{map}}^{\mathrm{ovl}}
=
-\log
\frac{
\frac{1}{|\mathcal S_{\mathrm{ovl}}|}
\sum_{i\in\mathcal S_{\mathrm{ovl}}}\bar{G}_i + \epsilon
}{
\frac{1}{|\mathcal S_{\mathrm{sel}}\setminus\mathcal S_{\mathrm{ovl}}|}
\sum_{i\in\mathcal S_{\mathrm{sel}}\setminus\mathcal S_{\mathrm{ovl}}}\bar{G}_i + \epsilon
}.
\end{equation}
The two map priors are combined as
\begin{equation}
\mathcal{L}_{\mathrm{map}}
=
\lambda_{\mathrm{sel}}\mathcal{L}_{\mathrm{map}}^{\mathrm{sel}}
+
\lambda_{\mathrm{ovl}}\mathcal{L}_{\mathrm{map}}^{\mathrm{ovl}}.
\end{equation}

\paragraph{Full objective.}
The full objective is
\begin{equation}
\mathcal{L}
=
\mathcal{L}_{\mathrm{main}}
+
\mathcal{L}_{\mathrm{sem}}
+
\mathcal{L}_{\mathrm{map}}.
\end{equation}
We set $\lambda_{cp}=0.01$, $\lambda_{off}=0.5$, 
$\lambda_{\mathrm{cue}}=\lambda_{\mathrm{map}}=0.1$, 
and $\lambda_{\mathrm{sel}}=\lambda_{\mathrm{ovl}}=10^{-4}$.

\begin{figure*}[t]
    \centering
    \setlength{\tabcolsep}{1pt}   
    \renewcommand{\arraystretch}{1.0}

    \newlength{\imgH}
    \setlength{\imgH}{0.82cm}     

    \newcommand{\method}[1]{%
        \parbox[c][\imgH][c]{1.55cm}{\centering\bfseries #1}%
    }

    \newcommand{\qimg}[1]{%
        \includegraphics[height=\imgH]{#1}%
    }

    \resizebox{\textwidth}{!}{%
    \begin{tabular}{@{}c*{20}{c}@{}}

        \method{Source} &
        \qimg{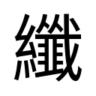} &
        \qimg{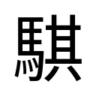} &
        \qimg{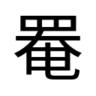} &
        \qimg{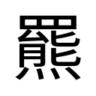} &
        \qimg{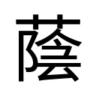} &
        \qimg{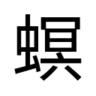} &
        \qimg{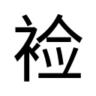} &
        \qimg{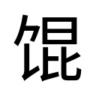} &
        \qimg{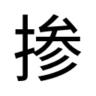} &
        \qimg{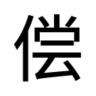} &
        \qimg{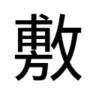} &
        \qimg{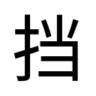} &
        \qimg{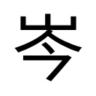} &
        \qimg{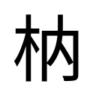} &
        \qimg{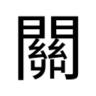} &
        \qimg{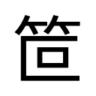} &
        \qimg{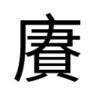} &
        \qimg{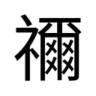} &
        \qimg{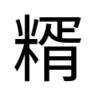} &
        \qimg{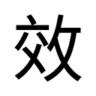} \\

        \method{MX-Font} &
        \qimg{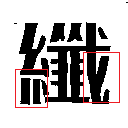} &
        \qimg{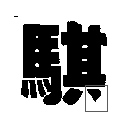} &
        \qimg{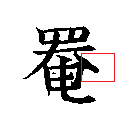} &
        \qimg{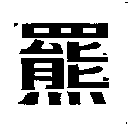} &
        \qimg{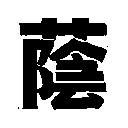} &
        \qimg{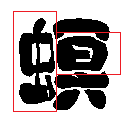} &
        \qimg{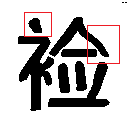} &
        \qimg{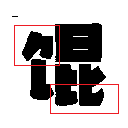} &
        \qimg{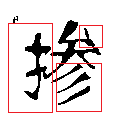} &
        \qimg{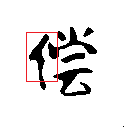} &
        \qimg{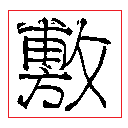} &
        \qimg{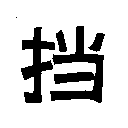} &
        \qimg{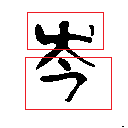} &
        \qimg{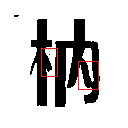} &
        \qimg{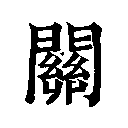} &
        \qimg{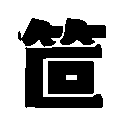} &
        \qimg{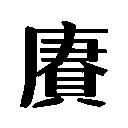} &
        \qimg{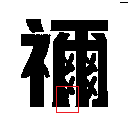} &
        \qimg{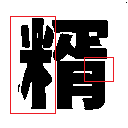} &
        \qimg{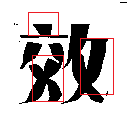} \\
        
        \method{Font Diffuser} &
        \qimg{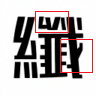} &
        \qimg{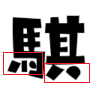} &
        \qimg{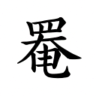} &
        \qimg{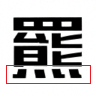} &
        \qimg{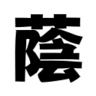} &
        \qimg{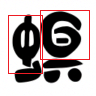} &
        \qimg{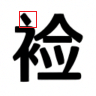} &
        \qimg{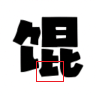} &
        \qimg{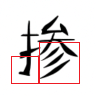} &
        \qimg{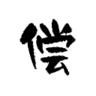} &
        \qimg{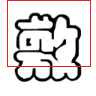} &
        \qimg{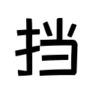} &
        \qimg{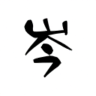} &
        \qimg{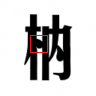} &
        \qimg{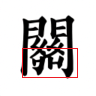} &
        \qimg{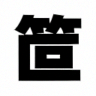} &
        \qimg{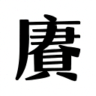} &
        \qimg{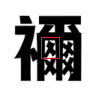} &
        \qimg{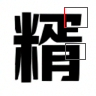} &
        \qimg{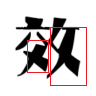} \\

        \method{CF-Font} &
        \qimg{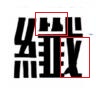} &
        \qimg{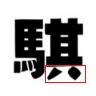} &
        \qimg{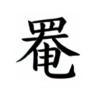} &
        \qimg{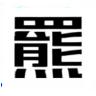} &
        \qimg{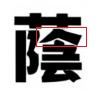} &
        \qimg{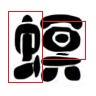} &
        \qimg{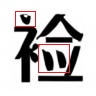} &
        \qimg{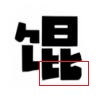} &
        \qimg{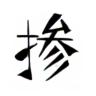} &
        \qimg{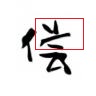} &
        \qimg{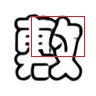} &
        \qimg{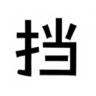} &
        \qimg{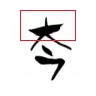} &
        \qimg{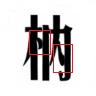} &
        \qimg{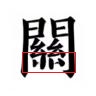} &
        \qimg{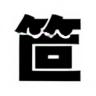} &
        \qimg{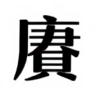} &
        \qimg{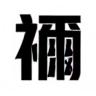} &
        \qimg{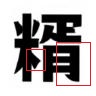} &
        \qimg{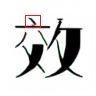} \\
        
        \method{VQ-Font} &
        \qimg{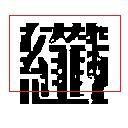} &
        \qimg{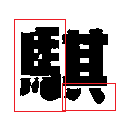} &
        \qimg{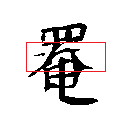} &
        \qimg{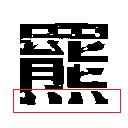} &
        \qimg{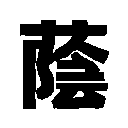} &
        \qimg{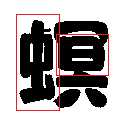} &
        \qimg{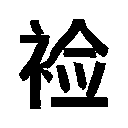} &
        \qimg{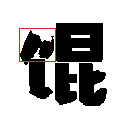} &
        \qimg{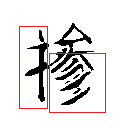} &
        \qimg{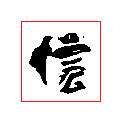} &
        \qimg{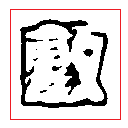} &
        \qimg{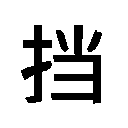} &
        \qimg{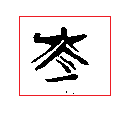} &
        \qimg{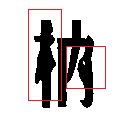} &
        \qimg{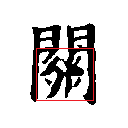} &
        \qimg{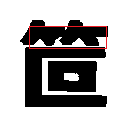} &
        \qimg{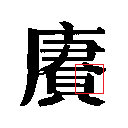} &
        \qimg{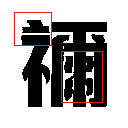} &
        \qimg{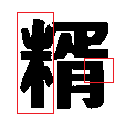} &
        \qimg{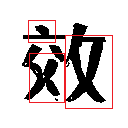} \\
        
        
        \method{Ours} &
        \qimg{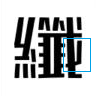} &
        \qimg{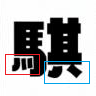} &
        \qimg{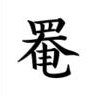} &
        \qimg{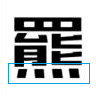} &
        \qimg{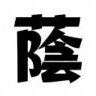} &
        \qimg{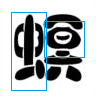} &
        \qimg{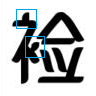} &
        \qimg{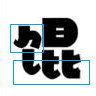} &
        \qimg{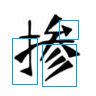} &
        \qimg{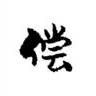} &
        \qimg{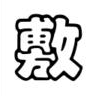} &
        \qimg{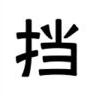} &
        \qimg{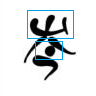} &
        \qimg{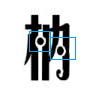} &
        \qimg{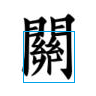} &
        \qimg{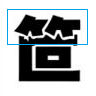} &
        \qimg{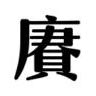} &
        \qimg{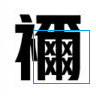} &
        \qimg{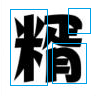} &
        \qimg{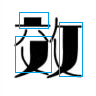}  \\

        \method{GT} &
        \qimg{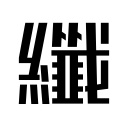} &
        \qimg{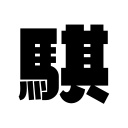} &
        \qimg{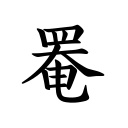} &
        \qimg{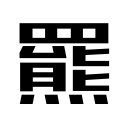} &
        \qimg{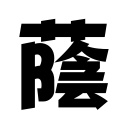} &
        \qimg{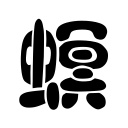} &
        \qimg{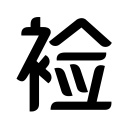} &
        \qimg{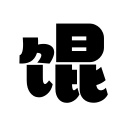} &
        \qimg{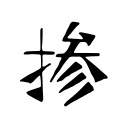} &
        \qimg{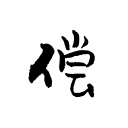} &
        \qimg{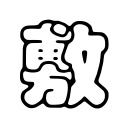} &
        \qimg{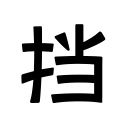} &
        \qimg{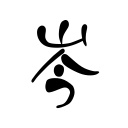} &
        \qimg{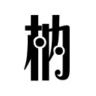} &
        \qimg{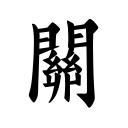} &
        \qimg{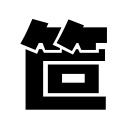} &
        \qimg{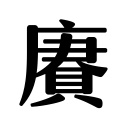} &
        \qimg{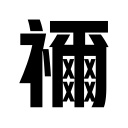} &
        \qimg{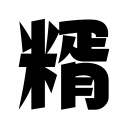} &
        \qimg{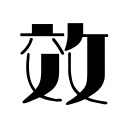}  \\

    \end{tabular}%
    }

    \caption{Qualitative comparison with representative few-shot font generation methods under the 3-shot setting.
    Red boxes mark regions with defective generation, while blue boxes highlight regions where our method produces more complete structures, better local stroke details, and stronger style consistency.
    }
    \label{fig:qualitative_grid}
\end{figure*}

\section{Experiments}

\subsection{Training Settings}

We construct the dataset from publicly available TTF fonts collected online. Each font is rendered into grayscale glyph images of size $96\times96$, yielding 291 training fonts and 30 test fonts. Under the few-shot setting, we use 3 reference glyphs for each target glyph in both training and inference. To increase the chance of informative local correspondence, we adopt an overlap-aware reference sampling strategy: with probability 50\%, at least one of the three references shares component overlap with the target glyph; otherwise, all references are sampled randomly.

The model is trained end-to-end from scratch, jointly optimizing the global diffusion backbone, the weakly supervised local expert branch, and the denoising-state condition allocation module. The local branch uses $K=8$ experts. We use AdamW with an initial learning rate of $1\times10^{-4}$, and train for 440{,}000 steps with a total batch size of 24. A cosine annealing schedule with restarts is adopted, with the learning rate reduced to $1\times10^{-5}$ in the final 10\% of training. Following standard classifier-free guidance training, conditional inputs are dropped with probability 0.1. All experiments are conducted in bfloat16 on 3 NVIDIA RTX 4090 GPUs, and training takes about 7 days.

At inference, we follow the same 3-shot protocol by randomly sampling 3 reference glyphs from the target font. We use classifier-free guidance with a scale of 7.5, and adopt DPM-Solver++ with 20 denoising steps for sampling.

\subsection{Quantitative comparison }

We compare SmartFont with representative few-shot font generation baselines, including global content--style modeling methods, component/local-prior methods, and recent diffusion-based methods\cite{mx_font,fontdiffuser,cf_font,pan2023vqlocal,lf_font_tpami}, under the same 3-shot setting.
Following the protocol above, we report FID, SSIM, LPIPS, and L1 distance on three difficulty levels, namely Easy, Medium, and Hard, as summarized in Tab.~\ref{tab:main_results}.

Overall, SmartFont achieves the best overall performance across all three difficulty levels.
The improvements are consistent over both fidelity-oriented metrics and perceptual metrics, indicating that the proposed multi-level condition allocation design better balances global glyph completeness and fine-grained local style fidelity than existing baselines.
Notably, the advantage remains clear on the more challenging Medium and Hard subsets, suggesting that the proposed framework is more robust when the target glyph requires stronger coordination between global structure preservation and localized style correction.

\begin{table*}[t]
    \centering
    \small
    \setlength{\tabcolsep}{4pt}
    \renewcommand{\arraystretch}{1.1}

    \newcommand{\best}[1]{\textbf{#1}}
    \newcommand{\second}[1]{\underline{#1}}

    \begin{tabular}{lcccccccccccc}
        \toprule
        & \multicolumn{4}{c}{Easy}
        & \multicolumn{4}{c}{Medium}
        & \multicolumn{4}{c}{Hard} \\
        \cmidrule(lr){2-5}
        \cmidrule(lr){6-9}
        \cmidrule(lr){10-13}
        Model
        & FID$\downarrow$ & SSIM$\uparrow$ & LPIPS$\downarrow$ & L1$\downarrow$
        & FID$\downarrow$ & SSIM$\uparrow$ & LPIPS$\downarrow$ & L1$\downarrow$
        & FID$\downarrow$ & SSIM$\uparrow$ & LPIPS$\downarrow$ & L1$\downarrow$ \\
        \midrule

        MX-Font   & 62.297 & 0.657 & 0.194 & 0.139 & 83.279 & 0.624 & 0.230 & 0.147 & 97.245 & 0.551 &  0.269 & 0.178 \\
        FontDiffuser & 32.341 & 0.744 & 0.168 & 0.076 &37.956 & 0.694 & 0.193 & 0.094 & 43.969  & 0.628 & 0.238 & 0.120 \\
        CF-Font & 41.631 & 0.761 & 0.163 & 0.070 & 45.576 & 0.710 & 0.192 & 0.089 & 49.389 & 0.641 & 0.238 & 0.114 \\
        VQ-Font & 55.098 & 0.691 & 0.202  & 0.132 & 66.506  & 0.662 & 0.245 & 0.141 & 84.401 & 0.589 & 0.296 & 0.174 \\
        LF-Font  &  & 0. &  & & .02 & 0. & 0. & 0. & . & 0. & 0. & 0. \\
        Ours    & \best{17.570} & \best{0.790} & \best{0.135} & \best{0.059}
                 & \best{26.442} & \best{0.731} & \best{0.168} & \best{0.080}
                 & \best{35.503} & \best{0.646} & \best{0.218} & \best{0.112} \\

        \bottomrule
    \end{tabular}

    \caption{Quantitative comparison with representative few-shot font generation baselines under the 3-shot setting.
    We report FID, SSIM, LPIPS, and L1 on Easy, Medium, and Hard subsets.}
    \label{tab:main_results}
\end{table*}

\subsection{Qualitative comparison}

Fig.~\ref{fig:qualitative_grid} shows qualitative comparisons with several representative baselines.
Compared with prior methods, SmartFont produces glyphs that are visually more complete and more consistent with the target font style, while preserving clearer fine-grained local stroke characteristics.
In contrast, existing methods more often exhibit insufficient local stylization, unstable stroke rendering, or weaker coordination between local details and the overall glyph structure.

These visual results are consistent with the quantitative results in Tab.~\ref{tab:main_results}.
They support our main claim that few-shot font generation should not be treated as a purely static global fusion problem, but instead benefits from explicitly coordinating global content, global style, and local corrective cues during denoising.

\subsection{Ablations}

\subsubsection{Effect of Different Modules}
\label{sec:ablation_modules}

\begin{table}[t]
    \centering
    \small
    \setlength{\tabcolsep}{5pt}
    \renewcommand{\arraystretch}{1.1}

    \begin{tabular}{lcccc}
        \toprule
        Ablation & FID$\downarrow$ & SSIM$\uparrow$ & LPIPS$\downarrow$ & L1$\downarrow$ \\
        \midrule
        + Baseline & 38.255 & 0.682 & 0.204 & 0.099 \\
        + Expert Encoder & 35.367 & 0.686 & 0.201 & 0.096 \\
        + Spatial Allocation & 29.320 & 0.709 & 0.184 & 0.088 \\
        + Layer Allocation & 25.942 & 0.721 & 0.177 & 0.084 \\
        \bottomrule
    \end{tabular}

\caption{Ablation study of different modules. Starting from the baseline, progressively adding the expert encoder, semantic-spatial allocation, and layer allocation consistently improves overall performance.}
\label{tab:ablation_table}
\end{table}

\begin{figure}[t]
    \centering
    \setlength{\tabcolsep}{0pt}
    \renewcommand{\arraystretch}{1.0}

    \newlength{\abcolw}
    \setlength{\abcolw}{0.163\linewidth} 

    \newlength{\abheadh}
    \setlength{\abheadh}{1.2em} 

    \newlength{\abimgh}
    \setlength{\abimgh}{2.1cm}  

    \newcommand{\abhead}[1]{%
        \parbox[c][\abheadh][b]{\abcolw}{\centering\scriptsize\bfseries #1}%
    }

    \newcommand{\abimg}[1]{%
        \parbox[c][\abimgh][c]{\abcolw}{%
            \centering
            \includegraphics[width=0.92\abcolw,height=\abimgh,keepaspectratio]{#1}%
        }%
    }

    \begin{tabular}{@{}cccccc@{}}
        \abhead{Content} &
        \abhead{+ Baseline} &
        \abhead{\shortstack[c]{+ Expert \\ Encoder}} &
        \abhead{\shortstack[c]{+ Spatial \\ Allocation}} &
        \abhead{+ Layer Allocation} &
        \abhead{GT} \\[-2pt]

        \abimg{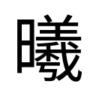} &
        \abimg{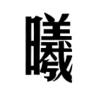} &
        \abimg{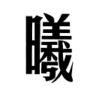} &
        \abimg{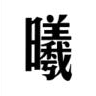} &
        \abimg{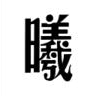} &
        \abimg{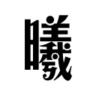}
    \end{tabular}
    \setlength{\abovecaptionskip}{1pt}
    \caption{Qualitative ablation results of different modules. As more proposed components are introduced, the generated glyph becomes progressively more faithful to the ground truth, especially in fine-grained local stroke details.}    \label{fig:ablation_fig}
\end{figure}

We progressively ablate the proposed modules to verify their individual contributions, including the local expert branch, the semantic-spatial allocation design, and the layer-level allocation mechanism. As shown in Tab.~\ref{tab:ablation_table}, simply introducing the expert encoder with naive spatial gating brings only marginal improvement over the baseline. In contrast, incorporating the semantic-spatial allocation yields a much clearer gain across all metrics, and further adding the layer-level allocation achieves the best overall performance.

The above results suggest that increasing local modeling capacity alone is insufficient. Without content-aware spatial map prediction and component-aware supervision, the expert branch cannot reliably learn where localized correction should be applied, and thus provides only limited benefit. In practice, such naive spatial gating tends to degenerate into trivial activation patterns, causing the expert branch to behave more like a stronger but still largely global style encoder rather than a semantically specialized local corrector. We defer detailed visualization and discussion of this phenomenon to Sec.~\ref{sec:vis_spatial}.

The qualitative results in Fig.~\ref{fig:ablation_fig} further support this observation. On this character, the improvements brought by the proposed modules are mainly reflected in fine-grained local stroke rendering rather than only coarse global structure. In particular, the full model better recovers subtle stroke details, such as the small dot-like stroke parts, while preserving the overall glyph layout. This indicates that the semantic-spatial allocation helps the model place local corrections on more appropriate regions, and the subsequent layer-level allocation further improves the coordination between global and local conditions during denoising.

\subsubsection{Visualization of Spatial-level Allocation}
\label{sec:vis_spatial}

\begin{figure}[t]
    \centering
    \setlength{\tabcolsep}{1.2pt}
    \renewcommand{\arraystretch}{1.0}

    \newlength{\expcolw}
    \setlength{\expcolw}{1.55cm}   

    \newlength{\expheadh}
    \setlength{\expheadh}{1.7em}

    \newlength{\expimgh}
    \setlength{\expimgh}{1.9cm}

    \newcommand{\exphead}[1]{%
        \parbox[c][\expheadh][b]{\expcolw}{\centering\scriptsize\bfseries #1}%
    }

    \newcommand{\expimg}[1]{%
        \parbox[c][\expimgh][c]{\expcolw}{%
            \centering
            \includegraphics[width=0.92\expcolw,height=\expimgh,keepaspectratio]{#1}%
        }%
    }

    \newcommand{\expimgred}[1]{%
        \parbox[c][\expimgh][c]{\expcolw}{%
            \centering
            {\setlength{\fboxsep}{0pt}%
             \setlength{\fboxrule}{1.1pt}%
             \fcolorbox{red}{white}{%
                \includegraphics[width=0.88\expcolw,height=\expimgh,keepaspectratio]{#1}%
             }}%
        }%
    }

    \newcommand{\expimgblk}[1]{%
        \parbox[c][\expimgh][c]{\expcolw}{%
            \centering
            {\setlength{\fboxsep}{0pt}%
             \setlength{\fboxrule}{1.4pt}%
             \fcolorbox{black}{white}{%
                \includegraphics[width=0.88\expcolw,height=\expimgh,keepaspectratio]{#1}%
             }}%
        }%
    }

    \resizebox{\linewidth}{!}{%
    \begin{tabular}{@{}cccccccc@{}}
        \exphead{Source} &
        \exphead{Ref} &
        \exphead{Expert 1} &
        \exphead{Expert 2} &
        \exphead{Expert 3} &
        \exphead{Expert 4} &
        \exphead{Expert 5} &
        \exphead{Expert 6} \\[-2pt]

        \expimg{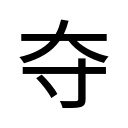} &
        \expimg{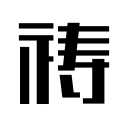} &
        \expimgred{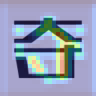} &
        \expimg{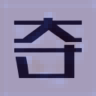} &
        \expimgblk{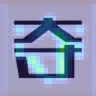} &
        \expimgblk{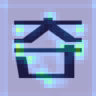} &
        \expimgblk{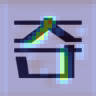} &
        \expimg{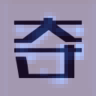} \\

        \expimg{./imgs/raw/ablation_heatmap/1/source.jpg} &
        \expimg{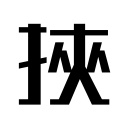} &
        \expimg{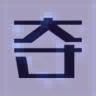} &
        \expimg{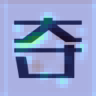} &
        \expimgblk{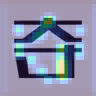} &
        \expimgblk{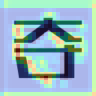} &
        \expimg{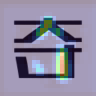} &
        \expimgred{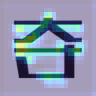} \\

        \expimg{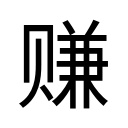} &
        \expimg{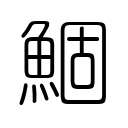} &
        \expimgblk{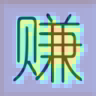} &
        \expimg{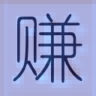} &
        \expimg{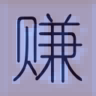} &
        \expimg{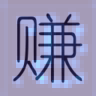} &
        \expimgblk{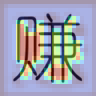} &
        \expimg{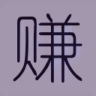} \\
    \end{tabular}%
    }

    \setlength{\abovecaptionskip}{3pt}
    \caption{Visualization of expert-wise outputs under different assignment conditions. 
    Black boxes indicate the experts selected by Hungarian-based component assignment, while red boxes denote the subset whose matched style-side components also overlap with the component set of the source glyph. 
    }
    \label{fig:expert_vis}
\end{figure}

\begin{figure}[t]
    \centering
    \makebox[\linewidth][c]{%
        \begin{subfigure}[t]{0.26\linewidth}
            \centering
            \includegraphics[width=\linewidth]{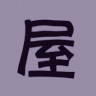}
            \caption{Inactive.}
        \end{subfigure}
        \hspace{0.02\linewidth}
        \begin{subfigure}[t]{0.26\linewidth}
            \centering
            \includegraphics[width=\linewidth]{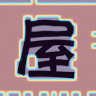}
            \caption{Background-focused.}
        \end{subfigure}
        \hspace{0.02\linewidth}
        \begin{subfigure}[t]{0.26\linewidth}
            \centering
            \includegraphics[width=\linewidth]{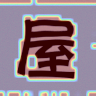}
            \caption{Full-glyph.}
        \end{subfigure}
    }
    \caption{Typical degeneration patterns of naive spatial gating without content-aware map prediction and component-aware supervision. The gate may become nearly inactive, shift its response to background regions, or cover the whole glyph, which weakens expert specialization and reduces the benefit of localized correction.}
    \label{fig:naive_gate_failure}
\end{figure}

We visualize the learned semantic-spatial allocation to examine how different experts respond under different assignment conditions. In Fig.~\ref{fig:expert_vis}, the first two rows share the same content glyph but use different reference glyphs, while the third row shows a non-overlap case where the assigned style-side component does not appear in the content glyph. 

Overall, the visualizations show a clear contrast between assigned and unassigned experts. Unassigned experts are largely suppressed, although weak residual responses may remain. In contrast, assigned experts consistently attend to meaningful local regions. When component overlap exists, their maps become more concentrated around the corresponding local part; when overlap does not exist, they still respond to structurally relevant regions rather than collapsing to arbitrary patterns. This suggests that the learned experts capture transferable local semantic cues instead of merely memorizing fixed component templates.

These results are consistent with the intended role of semantic-spatial allocation: the assignment cue determines which local concept an expert should model, while the learned spatial map determines where its corrective signal should be injected. By contrast, Fig.~\ref{fig:naive_gate_failure} shows that naive spatial gating often degenerates into inactive, background-focused, or full-glyph responses, which helps explain its limited gain in the ablation study.

\begin{figure}[t]
    \centering

    \begin{subfigure}[t]{\linewidth}
        \centering
        \includegraphics[page=10,width=\linewidth]{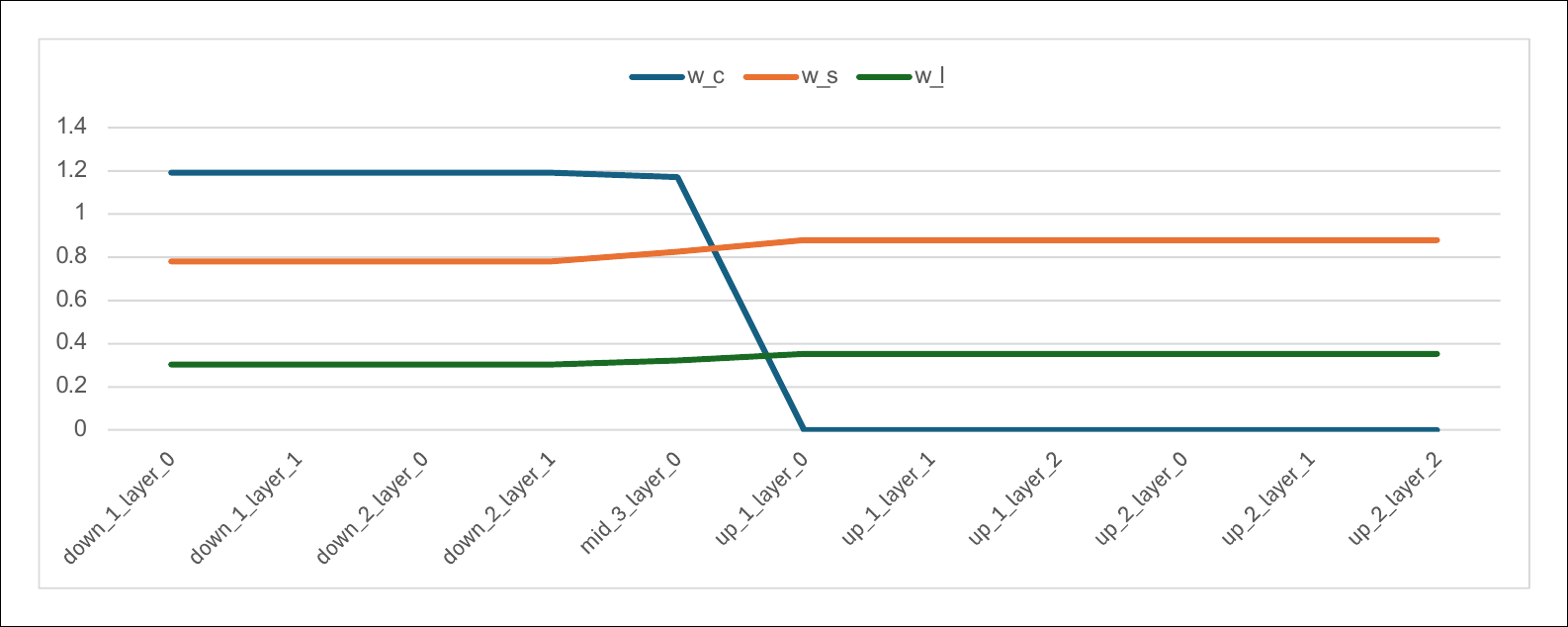}
        \caption{Base weight.}
        \label{fig:baseweight}
    \end{subfigure}

    \vspace{0.5em}

    \begin{subfigure}[t]{\linewidth}
        \centering
        \includegraphics[page=11,width=\linewidth]{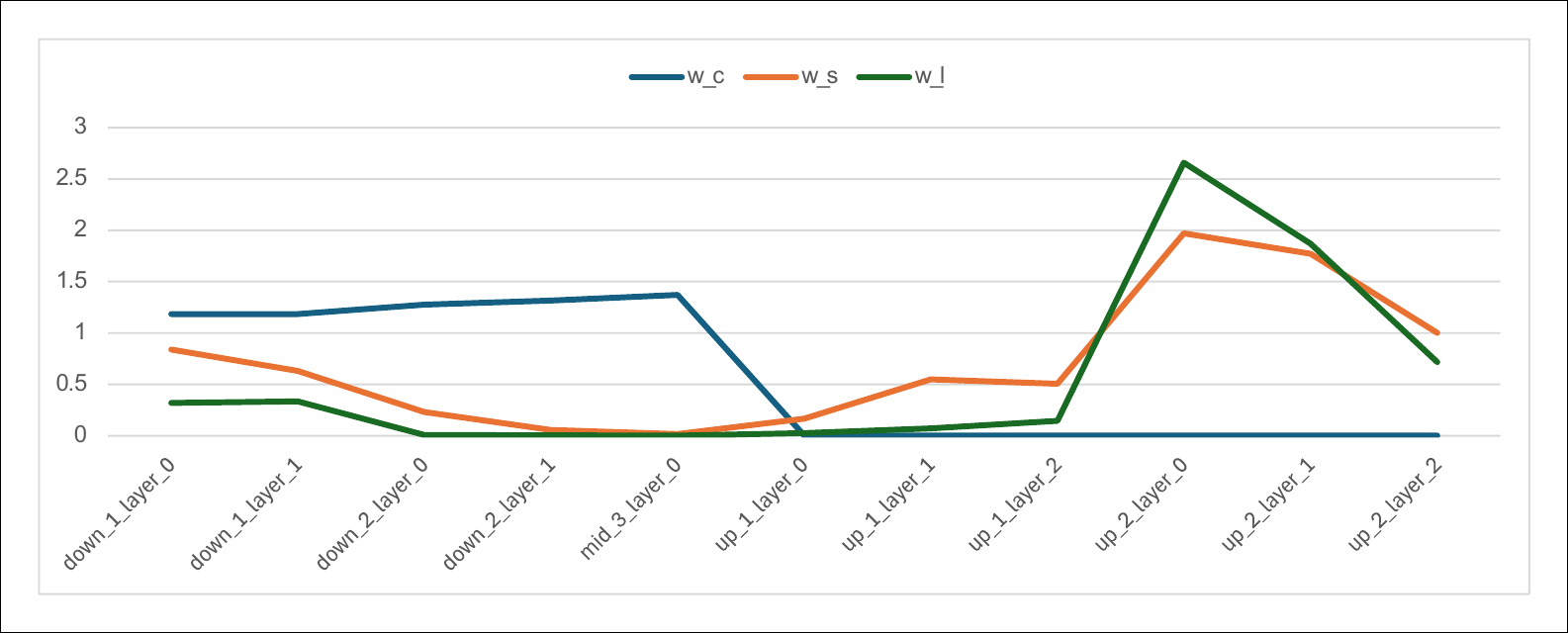}
        \caption{Learned weight.}
        \label{fig:learnedweight}
    \end{subfigure}

\caption{Comparison between the empirical base weights and the final learned weights at a representative denoising step ($t=19$). 
The three curves denote the coefficients for global content, global style, and local corrective residuals, respectively. 
The upper plot(a) shows the hand-crafted base schedule, while the lower plot(b) shows the refined weights after temporal and block-adaptive allocation.
}
\label{fig:weight_compare}
\end{figure}

\subsubsection{Visualization of Layer-level Allocation}
\label{sec:vis_layer}

We visualize the learned layer-level allocation in Fig.~\ref{fig:weight_compare} by comparing the empirical base weights with the final learned weights at denoising step $t=19$. 

The learned weights differ markedly from the empirical schedule rather than merely making small corrections. In the shown case, the final allocation becomes much more non-uniform across injection blocks, with several later blocks assigning stronger weights to style and local correction while suppressing the content pathway. Similar deviations are also observed across different inputs and timesteps: although the exact values vary, the learned allocation typically preserves only the broad tendency of the empirical prior while performing substantial state-dependent rebalancing. This suggests that a fixed manually specified weighting strategy is insufficient, and supports the need for the proposed learnable layer-level allocation mechanism.

\section{Conclusion}
In this paper, we revisited few-shot font generation from the perspective of multi-level condition allocation. Rather than treating the task as a purely static content--style fusion problem, we argued that its main difficulty lies in how imperfect yet complementary global and local conditions should be organized during generation. Based on this view, we proposed SmartFont, a diffusion-based framework that combines a global content--style backbone, a weakly supervised local expert branch for semantic-spatial allocation, and a denoising-state allocation module for adaptive condition weighting across timesteps and injection blocks. Extensive experiments and visualizations show that SmartFont achieves a better balance between global glyph completeness and fine-grained local style fidelity, while also providing interpretable evidence for both spatial-level and layer-level allocation behaviors. We hope this work offers a useful perspective for future few-shot font generation research, suggesting that adaptive condition coordination is as important as stronger representation learning itself.


\end{document}